\documentclass[10pt,twocolumn,letterpaper]{article}

\usepackage{cvpr}
\usepackage{times}
\usepackage{epsfig}
\usepackage{graphicx}
\usepackage{amsmath}
\usepackage{amssymb}
\usepackage{bbding}
\newcommand{\tabincell}[2]{\begin{tabular}{@{}#1@{}}#2\end{tabular}}
\usepackage{verbatim}
\usepackage{ulem}

\usepackage[breaklinks=true,bookmarks=false]{hyperref}
\cvprfinalcopy 

\def\cvprPaperID{2403} 

\ifcvprfinal\fi
\begin{document}

\title{Learning for Disparity Estimation through Feature Constancy}
\author{Zhengfa Liang\thanks{Equal contribution.} \quad Yiliu Feng$^{*}$  \quad Yulan Guo \quad Hengzhu Liu \quad Wei Chen\\Linbo Qiao\quad Li Zhou \quad Jianfeng Zhang \vspace{8pt}\\
National University of Defense Technology\\
{\tt\small \{liangzhengfa10, fengyiliu11, hengzhuliu, yulan.guo, chenwei,}\\
{\tt\small qiao.linbo, zhouli06, jianfengzhang\}@nudt.edu.cn}
}


\maketitle

\begin{abstract}
Stereo matching algorithms usually consist of four steps, including matching cost calculation, matching cost aggregation, disparity calculation, and disparity refinement. Existing CNN-based methods only adopt CNN to solve parts of the four steps, or use different networks to deal with different steps, making them difficult to obtain the overall optimal solution. In this paper, we propose a network architecture to incorporate all steps of stereo matching. The network consists of three parts. The first part calculates the multi-scale shared features.
The second part performs matching cost calculation, matching cost aggregation and disparity calculation to estimate the initial disparity using shared features.
The initial disparity and the shared features are used to calculate the feature constancy that measures correctness of the correspondence between two input images. The initial disparity and the feature constancy are then fed to a sub-network to refine the initial disparity.
The proposed method has been evaluated on the Scene Flow and KITTI datasets. It achieves the state-of-the-art performance on the KITTI 2012 and KITTI 2015 benchmarks while maintaining a very fast running time.
\end{abstract}

\section{Introduction}

Stereo matching aims to estimate correspondences of all pixels between two rectified images \cite{Barnard1982Computational,Scharstein2002A,Hartley2003Multiple}. It is a core problem for many stereo vision tasks and has numerous applications in areas such as autonomous vehicles \cite{Sivaraman2013ARO}, robotics navigation \cite{Schmid2013StereoVB}, and augmented reality \cite{Zenati2008Dense}.

Stereo matching has been intensively investigated for several decades, with a popular four-step pipeline being developed. This pipeline includes matching cost calculation, matching cost aggregation, disparity calculation and disparity refinement \cite{Scharstein2002A}. The four-step pipeline dominants existing stereo matching algorithms \cite{Scharstein2003High, Scharstein2007Learning, Hirschmuller2007Evaluation, Scharstein2014High}, while each of its steps is important to the overall stereo matching performance. Due to the powerful representative capability of deep convolution neural network (CNN) for various vision tasks \cite{Krizhevsky2012ImageNet,Wang2013Learning,Long2015Fully}, CNN has been employed to improve stereo matching performance and outperforms traditional methods significantly \cite{Shaked2016Improved,zbontar2016stereo,Kendall2017End,Mayer2016A,Luo2016Efficient}.

Zbontar and LeCun \cite{zbontar2016stereo} first introduced CNN to calculate the matching cost to measure the similarity of two pixels of two images. This method achieved the best performance on the KITTI 2012 \cite{Geiger2012Are}, KITTI 2015 \cite{Menze2015Object} and Middlebury \cite{Scharstein2002A, Scharstein2003High, Scharstein2007Learning, Hirschmuller2007Evaluation, Scharstein2014High} stereo datasets at that time. They argued that it is unreliable to consider only the difference of photometry in pixels or hand-crafted image features for matching cost. In contrast, CNN can learn more robust and discriminative features from images, and produces improved stereo matching cost.
Following the work \cite{zbontar2016stereo}, several methods were proposed to improve the computational efficiency \cite{Luo2016Efficient} or matching accuracy \cite{Shaked2016Improved}. However, these methods still suffer from few limitations. \textbf{First}, to calculate the matching cost at all potential disparities, multiple forward passes have to be conducted by the network, resulting in high computational burden. \textbf{Second}, the pixels in occluded regions (i.e., only visible in one of the two images) cannot be used to perform training. It is therefore difficult to obtain a reliable disparity estimation in these regions. \textbf{Third}, several heuristic post-processing steps are required to refine the disparity. The performance and the generalization ability of these methods are therefore limited, as a number of parameters have to be chosen empirically.

Alternatively, the matching cost calculation, matching cost aggregation and disparity calculation steps can be seamlessly integrated into a CNN to directly estimate the disparity from stereo images \cite{Mayer2016A,Kendall2017End}. Traditionally, the matching cost aggregation and disparity calculation steps are solved by minimizing an energy function defined upon matching costs. For example, the Semi-Global Matching (SGM) method \cite{Hirschm2008Stereo}  uses dynamic programming to optimize a path-wise form of the energy function in several directions. Both the energy function and its solving process are hand-engineered. Different from the traditional methods, Mayer et al. \cite{Mayer2016A} and Kendall et al. \cite{Kendall2017End} directly stacked several convolution layers upon the matching costs, and trained the whole neural network to minimize the distance between the network output and the groundtruth disparity. These methods achieve higher accuracy and computational efficiency than the methods that use CNN for matching cost calculation only.

If all steps are integrated into a whole network for joint optimization, better disparity estimation performance can be expected. However, it is non-trivial to integrate the disparity refinement step with the other three steps. Existing methods \cite{Gidaris2017Detect, Pang2017Cascade} used additional networks for disparity refinement. Specifically,  once the disparity is calculated by CNN, one network or multiple networks are introduced to model the joint space of the inputs (including stereo images and initial disparity) and the output (i.e., refined disparity) to refine the disparity.

To bridge the gap between disparity calculation and disparity refinement, we propose to use feature constancy to identify the correctness of the initial disparity, and \textcolor{black}{then perform disparity refinement using feature constancy}. Here, ``constancy'' is borrowed from the area of optical estimation, where ``grey value constancy'' and ``gradient constancy'' are used \cite{Brox2004High}. ``Feature constancy'' means the correspondence of two pixels in feature space. Specifically, \textcolor{black}{the feature constancy includes two terms, i.e., feature correlation and reconstruction error. The correlation between features extracted from left and right images is considered as the first feature constancy term, which measures the correspondence at all possible disparities. 
The reconstruction error in feature space is considered as the second feature constancy term estimated with the knowledge on initial disparity. Then, the disparity refinement task \textcolor{black}{aims to improve the quality of the initial disparity given the feature constancy}, this can be implemented by a small sub-network.  These will be further explained in Sec. \ref{sec:disparity-refinement-subnetwork}. }Experiments on the Scene Flow and KITTI datasets have showed the effectiveness of our disparity refinement approach. Our method seamlessly integrates the disparity calculation and disparity refinement into one network for joint optimization, and improves the accuracy of the initial disparity by a notable margin. Our method achieves the state-of-the-art performance on the KITTI 2012 and KITTI 2015 benchmarks. It also has a very high running efficiency.

The contributions of this paper can be summarized as follows: 1) We integrate all steps of stereo matching into one network to improve accuracy and efficiency; 2) We \textcolor{black}{perform  disparity refinement with a sub-network} using the feature constancy; 3) We achieve the state-of-the-art performance on the KITTI benchmarks.

\section{Related works}
Over the last few years, CNN has been introduced to solve various problems in stereo matching. Existing CNN-based methods can broadly be divided into the following three categories.

\subsection{CNN for Matching Cost Learning}

In this category, CNN is used to learn the matching cost.  Zbontar and LeCun \cite{zbontar2016stereo} trained a CNN to compute the matching cost between two image patches (e.g., 9 $\times$ 9), which is followed by several post-processing steps, including cross-based cost aggregation, semi-global matching, left-right constancy check, sub-pixel enhancement, median filtering and bilateral filtering. This architecture needs multiple forward passes to calculate matching cost at all possible disparities. Therefore, this method is computationally expensive. Luo et al. \cite{Luo2016Efficient} introduced a product layer to compute the inner product between the two representations of a siamese architecture, and trained the network as multi-class classification over all possible disparities to reduce computational time. Park and Lee \cite{Park2017Look} introduced a pixel-wise pyramid pooling scheme to enlarge the receptive field during the comparison of two input patches. This method produced more accurate matching cost than \cite{zbontar2016stereo}. Shaked and Wolf \cite{Shaked2016Improved} deepened the network for matching cost calculation using a highway network architecture with multi-level weighted residual shortcuts. It was demonstrated that this architecture outperformed several networks, such as the base network from MC-CNN \cite{zbontar2016stereo}, the conventional high-way network \cite{Srivastava2015HighwayN}, ResNets \cite{He2016DeepRL}, DenseNet \cite{Huang2016Densely}, and the ResNets of ResNets \cite{Zhang2016Residual}.

\subsection{CNN for Disparity Regression}
In this category, CNN is carefully designed to directly estimate the disparity, which enables end-to-end training. Mayer et al. \cite{Mayer2016A} proposed an encoder-decoder architecture for disparity regression. The matching cost calculation is seamlessly integrated to the encoder part. The disparity is directly regressed in a forward pass. Kendall et al. \cite{Kendall2017End} used 3-D convolutions upon the matching costs to incorporate contextual information and introduced a differentiable ``soft argmin'' operation to regress the disparity. Both methods can run very fast, with 0.06s and 0.9s consumed on a single Nvidia GTX Titan X GPU, respectively.
However, disparity refinement is not included in these networks, which limits their performance.

\begin{figure*}[!htb]
\begin{center}
   \includegraphics[width=1\linewidth]{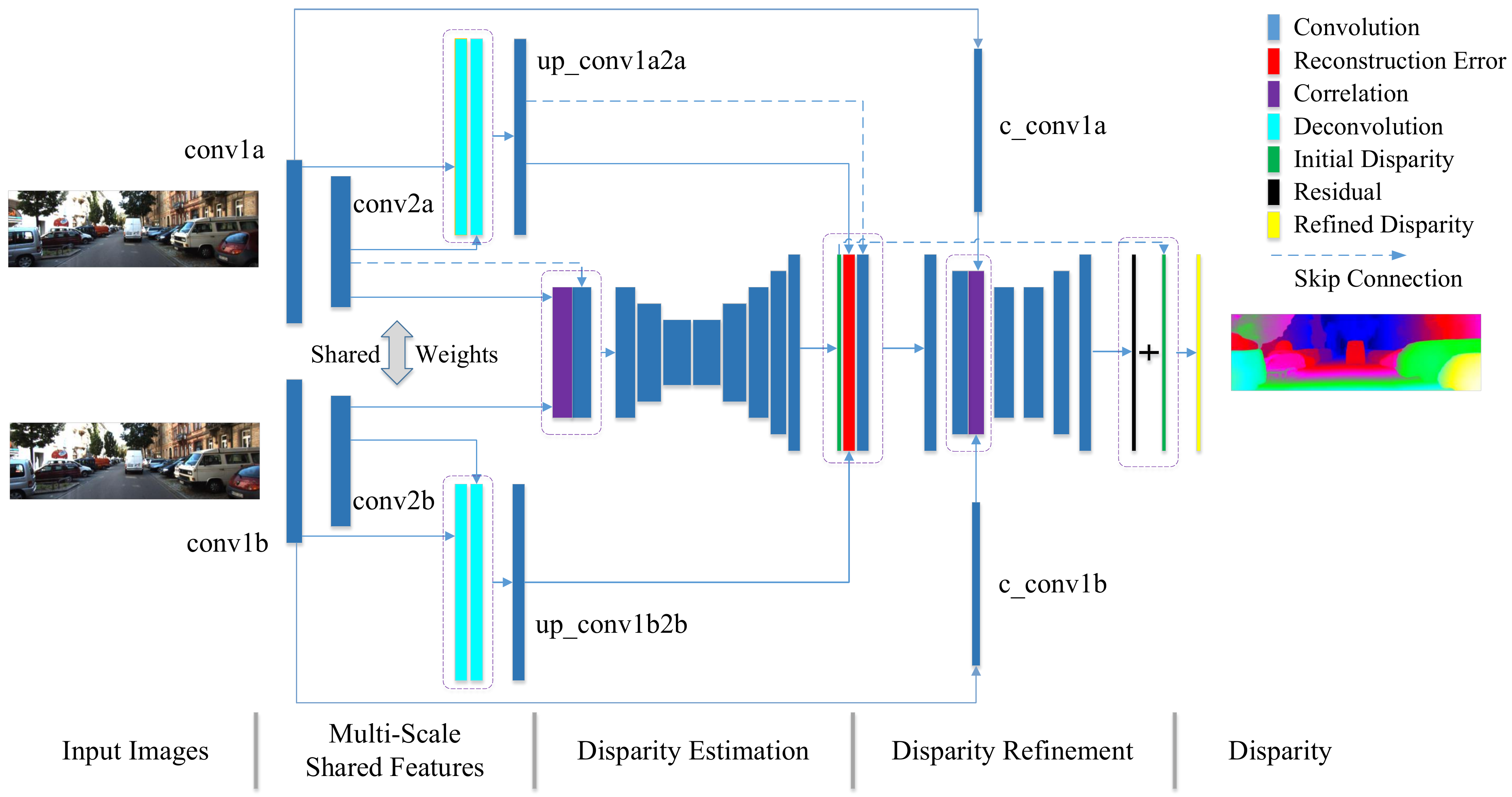}
\end{center}
   \caption{The architecture of our proposed network. It incorporates all of the four steps for stereo matching into a single network. \textcolor{black}{Note that the skip connections between encoder and decoder at different scales are omitted for better visualization.}}
\label{fig:long}
\label{fig:onecol}
\end{figure*}

\subsection{Multiple \textcolor{black}{Sub-networks}}
In this category, different networks are used to handle the four steps for stereo matching. Shaked and Wolf \cite{Shaked2016Improved} used their highway network for matching cost calculation and an additional global disparity CNN to replace the ``winner-takes-all'' strategy used in conventional matching cost aggregation and disparity calculation steps. This method improves performance in several challenging situations, such as in occluded, distorted, highly reflective and sparse textured regions. Gidaris et al. \cite{Gidaris2017Detect} used the method in \cite{Luo2016Efficient} to calculate the initial disparity, and then applied three additional neural networks for disparity refinement. Seki and Pollefeys \cite{SekiSGMNetsSM} proposed SGM-Nets to learn the SGM parametrization. They obtained better penalties than the hand tuned method used in MC-CNN \cite{zbontar2016stereo}. Peng et. al  \cite{Pang2017Cascade} built their work upon \cite{Mayer2016A} by cascading an additional network for disparity refinement.

In our work, we incorporate all steps into one network. As a result, all steps can share the same features and can be optimized jointly. \textcolor{black}{Besides, we introduce feature constancy into our network for improved disparity refinement using both feature correlation and reconstruction error}. It is clearly demonstrated that better disparity estimation performance can be achieved by our method.

\section{Approach}
\label{sec:app}

Different from existing methods that use multiple networks for different steps in stereo matching, we incorporate all step into a single network to enable end-to-end training.
The proposed network consists of three parts: multi-scale shared feature extraction, initial disparity estimation and disparity refinement. The framework of the proposed network is shown in Fig. \ref{fig:onecol}, and the network architecture is described in Table \ref{table:iresnet}.

\subsection{Stem Block for Multi-scale Feature Extraction}
\label{sec:stem_block}

The stem block extracts multi-scale shared features from the two input images for both initial disparity estimation and disparity refinement sub-networks. It contains two convolution layers with stride of 2 to reduce the resolution of inputs, and two deconvolution layers to up-sample the outputs of the two convolution layers to full-resolution. The up-sampled features are fused through an additional $1\times1$ convolution layer. An illustration is shown in Figure \ref{fig:onecol}. The outputs of this stem block can be divided into three types:
\begin{itemize}
\item[1)] The outputs of the second convolution layer (i.e., $conv2a$ for the left image and $conv2b$ for the right image). Correlation with a large displacement (i.e., 40) is performed between $conv2a$ and $conv2b$ to capture \textcolor{black}{the long-range but coarse-grained} correspondence between two images. It is used by the first sub-network for initial disparity estimation.
\item[2)] The outputs of the first convolution layer (i.e., $conv1a$ and $conv1b$). They are first compressed to fewer channels to obtain $c\_conv1a$ and $c\_conv1b$ through a convolution layer with a kernel size of 3$\times$3, on which correlation with a small displacement (i.e., 20) is performed to capture \textcolor{black}{short-range but fine-grained} correspondence, which is complementary to the former \textcolor{black}{correlation}. Besides, these features also act as the \textcolor{black}{\textbf{first feature constancy term}} used by the second sub-network.
\item[3)] Multi-scale fusion features (i.e., $up\_1a2a$ and $up\_1b2b$). They are first used as skip connection features to bring detailed information for the first sub-network. They are then used to calculate the \textcolor{black}{\textbf{second feature constancy term}} for the second sub-network.
\end{itemize}

\subsection{Initial Disparity Estimation Sub-network\label{sec:Disparity-Estimation-Sub-network}}

This sub-network generates a disparity map from ``$conv2a$'' and ``$conv2b$'' through an encoder-decoder architecture\textcolor{black}{, which is inspired by DispNetCorr1D \cite{Mayer2016A}. DispNetCorr1D can only output disparity of half resolution. By using the full-resolution multi-scale fusion features as skip connection features, we are able to estimate initial disparity of full resolution. The multi-scale fusion features are also used to calculate the reconstruction error, as will be described in Sec. \ref{sec:disparity-refinement-subnetwork}. In this sub-network, a correlation layer is first} introduced to calculate the matching costs in feature space.
There is a trade-off between accuracy and computational efficiency for matching cost calculation. That is, if matching cost is calculated using high-level features, more details are lost and several similar correspondences cannot be distinguished. In contrast, if matching cost is calculated using low-level features, the computational cost is high as feature maps are too large, and the receptive field is too small to capture robust features.

The matching cost is then concatenated with features from the left image. By concatenation, we expect the sub-network to consider low-level semantic information when performing disparity estimation over the matching costs. This to some extend \textcolor{black}{help} aggregate the matching cost and improves disparity estimation.

Disparity estimation is performed in the decoder part at different scales, where skip connection is introduced at each scale, as illustrated in Table \ref{table:iresnet}. For the sake of computational efficiency, the multi-scale fusion features (described in Sec. \ref{sec:stem_block}) are only skip connected to the last layer of the decoder to perform full-resolution disparity estimation. \textcolor{black}{This sub-network is called \textbf{DES-net}}.



\begin{table}[!h]
\caption{The detailed architecture of our network. \textcolor{black}{Note that in the ``Input'' column, ``+'' means concatenation and is fed into one bottom blob, while ``,'' means that the inputs are fed into different bottom blobs (the input channel number means the channel number of each bottom blob in this case).}}
\label{table:iresnet}
\centering
\scriptsize
\begin{tabular}{l|l|l|l|l|l}
\hline
Type& Name &\multicolumn{1}{c}{k} & \multicolumn{1}{c}{s} & \multicolumn{1}{l}{c I/O}  & Input \\
\hline
\multicolumn{6}{c}{\textit{Stem Block for Multi-scale Shared Features Extraction }}\\
\hline
Conv&\tabincell{l}{conv1a\\conv1b}&     \multicolumn{1}{c}{7}  & \multicolumn{1}{c}{2} & \multicolumn{1}{l}{3/64}    & \tabincell{l}{left image \\ right image} \\
Deconv&\tabincell{l}{up\_1a\\up\_1b}     &\multicolumn{1}{c}{4} & \multicolumn{1}{c}{2} & \multicolumn{1}{l}{64/32}     &  \tabincell{l}{conv1a\\conv1b} \\
Conv&\tabincell{l}{conv2a\\conv2b}&     \multicolumn{1}{c}{5}  & \multicolumn{1}{c}{2} & \multicolumn{1}{l}{64/128}  & \tabincell{l}{conv1a\\conv1b} \\
Deconv&\tabincell{l}{up\_2a\\up\_2b}  &\multicolumn{1}{c}{8} & \multicolumn{1}{c}{4} & \multicolumn{1}{l}{128/32}    &  \tabincell{l}{conv2a\\conv2b}\\
Conv&\tabincell{l}{up\_1a2a\\ up\_1b2b} &\multicolumn{1}{c}{1} & \multicolumn{1}{c}{1} & \multicolumn{1}{l}{64/32}   &  \tabincell{l}{up\_1a+up\_2a\\ up\_1b+up\_2b}\\
\hline
\multicolumn{6}{c}{\textit{Initial Disparity Estimation Sub-network}}\\
\hline
Corr&corr1d&\multicolumn{1}{c}{1} & \multicolumn{1}{c}{1} & \multicolumn{1}{l}{128/81} &  conv2a, conv2b \\
Conv&conv\_redir&\multicolumn{1}{c}{1} & \multicolumn{1}{c}{1} & \multicolumn{1}{l}{128/64}   & conv2a \\
Conv&conv3&\multicolumn{1}{c}{3}       & \multicolumn{1}{c}{2} & \multicolumn{1}{l}{145/256}  & corr1d+conv\_redir \\
Conv&conv3\_1&\multicolumn{1}{c}{3}    & \multicolumn{1}{c}{1} & \multicolumn{1}{l}{256/256}  & conv3 \\
Conv&conv4&\multicolumn{1}{c}{3}       & \multicolumn{1}{c}{2} & \multicolumn{1}{l}{256/512}  & conv3\_1 \\
Conv&conv4\_1&\multicolumn{1}{c}{3}    & \multicolumn{1}{c}{1} & \multicolumn{1}{l}{512/512}  & conv4 \\
Conv&conv5&\multicolumn{1}{c}{3}       & \multicolumn{1}{c}{2} & \multicolumn{1}{l}{512/512}  & conv4\_1 \\
Conv&conv5\_1&\multicolumn{1}{c}{3}    & \multicolumn{1}{c}{1} & \multicolumn{1}{l}{512/512}  & conv5 \\
Conv&conv6&\multicolumn{1}{c}{3}       & \multicolumn{1}{c}{2} & \multicolumn{1}{l}{512/1024} & conv5\_1 \\
Conv&conv6\_1&\multicolumn{1}{c}{3}    & \multicolumn{1}{c}{1} & \multicolumn{1}{l}{1024/1024}& conv6 \\
Conv&disp6 &\multicolumn{1}{c}{3} & \multicolumn{1}{c}{1} & \multicolumn{1}{l}{1024/1}    &  conv6\_1 \\
Deconv&uconv5     &\multicolumn{1}{c}{4} & \multicolumn{1}{c}{2} & \multicolumn{1}{l}{1024/512}  &  conv6\_1 \\
Conv&iconv5      &\multicolumn{1}{c}{3} & \multicolumn{1}{c}{1} & \multicolumn{1}{l}{1025/512}  &  uconv5+disp6+conv5\_1 \\
Conv&disp5 &\multicolumn{1}{c}{3} & \multicolumn{1}{c}{1} & \multicolumn{1}{l}{512/1}     &  iconv5 \\
Deconv&uconv4     &\multicolumn{1}{c}{4} & \multicolumn{1}{c}{2} & \multicolumn{1}{l}{512/256}   &  iconv5 \\
Conv&iconv4      &\multicolumn{1}{c}{3} & \multicolumn{1}{c}{1} & \multicolumn{1}{l}{769/256}   &  uconv4+disp5+conv4\_1 \\
Conv&disp4 &\multicolumn{1}{c}{3} & \multicolumn{1}{c}{1} & \multicolumn{1}{l}{256/1}     &  iconv4 \\
Deconv&uconv3     &\multicolumn{1}{c}{4} & \multicolumn{1}{c}{2} & \multicolumn{1}{l}{256/128}   &  iconv4 \\
Conv&iconv3      &\multicolumn{1}{c}{3} & \multicolumn{1}{c}{1} & \multicolumn{1}{l}{385/128}   &  uconv3+disp4+conv3\_1 \\
Conv&disp3 &\multicolumn{1}{c}{3} & \multicolumn{1}{c}{1} & \multicolumn{1}{l}{128/1}     &  iconv3 \\
Deconv&uconv2     &\multicolumn{1}{c}{4} & \multicolumn{1}{c}{2} & \multicolumn{1}{l}{128/64}    &  iconv3 \\
Conv&iconv2      &\multicolumn{1}{c}{3} & \multicolumn{1}{c}{1} & \multicolumn{1}{l}{193/64}    &  uconv2+disp3+conv2a \\
Conv&disp2 &\multicolumn{1}{c}{3} & \multicolumn{1}{c}{1} & \multicolumn{1}{l}{64/1}      &  iconv2 \\
Deconv&uconv1     &\multicolumn{1}{c}{4} & \multicolumn{1}{c}{2} & \multicolumn{1}{l}{64/32}     &  iconv2 \\
Conv&iconv1      &\multicolumn{1}{c}{3} & \multicolumn{1}{c}{1} & \multicolumn{1}{l}{97/32}     &  uconv1+disp2+conv1a  \\
Conv&disp1 &\multicolumn{1}{c}{3} & \multicolumn{1}{c}{1} & \multicolumn{1}{l}{32/1}      &  iconv1 \\
Deconv&uconv0      &\multicolumn{1}{c}{4} & \multicolumn{1}{c}{2} & \multicolumn{1}{l}{32/32}     &  iconv1 \\
Conv&iconv0       &\multicolumn{1}{c}{3} & \multicolumn{1}{c}{1} & \multicolumn{1}{l}{65/32}     &  \tabincell{c}{uconv0+disp1+up\_1a2a}  \\
Conv&disp0  &\multicolumn{1}{c}{3} & \multicolumn{1}{c}{1} & \multicolumn{1}{l}{32/1}      &  iconv0 \\
\hline
\multicolumn{6}{c}{\textit{Disparity Refinement Sub-network}}\\
\hline
Warp&w\_up\_1b2b     &\multicolumn{1}{c}{-} & \multicolumn{1}{c}{-} & \multicolumn{1}{l}{32/32}  &  up\_1b2b \\
Conv&r\_conv0    &\multicolumn{1}{c}{3}    & \multicolumn{1}{c}{1} & \multicolumn{1}{l}{65/32}  & \tabincell{l}{$\|$up\_1a2a-w\_up\_1b2b$\|$\\+disp0+up\_1a2a}\\
Conv&r\_conv1    &\multicolumn{1}{c}{3}    & \multicolumn{1}{c}{2} & \multicolumn{1}{l}{32/64}  & r\_conv0 \\
Conv&\tabincell{l}{c\_conv1a\\c\_conv1b}  &\multicolumn{1}{c}{3} & \multicolumn{1}{c}{1} & \multicolumn{1}{l}{64/16}    &  \tabincell{l}{conv1a\\conv1b}\\
Corr&r\_corr & \multicolumn{1}{c}{1}    & \multicolumn{1}{c}{1} & \multicolumn{1}{l}{16/41}  & c\_conv1a, c\_conv1b\\
Conv&r\_conv1\_1 &\multicolumn{1}{c}{3}    & \multicolumn{1}{c}{1} & \multicolumn{1}{l}{105/64}  & r\_conv1+r\_corr \\
Conv&r\_conv2    &\multicolumn{1}{c}{3}    & \multicolumn{1}{c}{2} & \multicolumn{1}{l}{64/128}  & r\_conv1\_1 \\
Conv&r\_conv2\_1 &\multicolumn{1}{c}{3}    & \multicolumn{1}{c}{1} & \multicolumn{1}{l}{128/128}  & r\_conv2 \\
Conv&r\_res2 &\multicolumn{1}{c}{3} & \multicolumn{1}{c}{1} & \multicolumn{1}{l}{128/1}   &  r\_conv2\_1 \\
Deconv&r\_uconv1     &\multicolumn{1}{c}{4} & \multicolumn{1}{c}{2} & \multicolumn{1}{l}{128/64}   &  r\_conv2\_1 \\
Conv&r\_iconv1      &\multicolumn{1}{c}{3} & \multicolumn{1}{c}{1} & \multicolumn{1}{l}{127/64}   &  r\_uconv1+r\_res2+r\_conv1\_1  \\
Conv&r\_res1 &\multicolumn{1}{c}{3} & \multicolumn{1}{c}{1} & \multicolumn{1}{l}{64/1}    &  r\_iconv1 \\
Deconv&r\_uconv0      &\multicolumn{1}{c}{4} & \multicolumn{1}{c}{2} & \multicolumn{1}{l}{64/32}   &  r\_iconv1 \\
Conv&r\_iconv0      &\multicolumn{1}{c}{3} & \multicolumn{1}{c}{1} & \multicolumn{1}{l}{65/32}    &  r\_uconv1+r\_res1+r\_conv0  \\
Conv&r\_res0 &\multicolumn{1}{c}{3} & \multicolumn{1}{c}{1} & \multicolumn{1}{l}{32/1}    &  r\_iconv0 \\
\hline
\end{tabular}
\end{table}

\subsection{Disparity Refinement Sub-network \label{sec:disparity-refinement-subnetwork}}

Although the disparity map estimated in Sec. \ref{sec:Disparity-Estimation-Sub-network} is already good, it still suffers from several challenges such as depth discontinuities and outliers. Consequently, disparity refinement is required to further improve the depth estimation performance.

In this paper, we \textcolor{black}{perform disparity refinement using feature constancy. Specifically, after obtaining the initial disparity $disp_i$ using DES-net, we calculate the two feature constancy terms (i.e., feature correlation $fc$ and reconstruction error $re$) . Then, the task of disparity refinement is to obtain the refined disparity $disp_r$ considering these three types of information, i.e.,
\begin{equation}
\small
\left\{ disp_i, fc, re \right\} \xrightarrow[]{CNN}   disp_r  \label{eq:disp-ftr}
\end{equation}\label{eq:bayes}\\
Specifically, the first feature constancy term $fc$ is calculated as the correlation between the feature maps of the left and right images (i.e., $c\_conv1a$ and $c\_conv1b$). $fc$ measures the correspondence of two feature maps at all displacements in disparity range that considered. It would produce large values at correct disparities.
The second feature constancy term $re$ is calculated as the reconstruction error of the initial disparity, i.e., the absolute difference between the multi-scale fusion features (Sec. \ref{sec:stem_block} ) of the left image and the back-warped features of the right image. Note that, to calculate $re$, only one displacement is conducted at each location in the feature maps, which relies on the corresponding value of the initial disparity. If the reconstruction error is large, the estimated disparity is more likely to be incorrect or from occluded regions.}

In practice, given the initial disparity produced by the disparity estimation sub-network (Sec. \ref{sec:Disparity-Estimation-Sub-network}), the disparity refinement sub-network estimates the residual to the initial disparity. The summation of the residual and the initial disparity is considered as the refined disparity map. Since both the initial disparity and the two feature constancy terms are used to produce the disparity map $disp_r$ (as shown in Eq. \ref{eq:disp-ftr}), the disparity estimation performance is expected to be improved. \textcolor{black}{This sub-network is called \textbf{DRS-net}}. Note that, since the four steps for stereo matching are integrated into a single CNN network, end-to-end training is ensured.

\subsection{Iterative Refinement}

To extract more information from the multi-scale fusion features and to ultimately improve the disparity estimation accuracy, an iterative refinement approach is proposed. Specifically, the refined disparity map produced by the second sub-network (Sec. \ref{sec:disparity-refinement-subnetwork}) is considered as a new initial disparity map, the feature constancy calculation and disparity refinement processes are then repeated to obtain a new refined disparity. This procedure is repeated until the improvement between two iterations is small. Note that, as the number of iterations is increased, the improvement decreases. 

\section{Experiments}

In this section, we evaluate our method iResNet (\textbf{i}terative \textbf{res}idual prediction \textbf{net}work) on two datasets, i.e., Scene Flow \cite{Mayer2016A} and KITTI \cite{Geiger2012Are, Menze2015Object}. The Scene Flow dataset \cite{Mayer2016A} is a synthesised dataset containing 35, 454 training image pairs and 4, 370 testing image pairs. Dense groundtruth disparities are provided for both training and testing sets. Besides, this dataset is sufficiently large to train a model without over-fitting. Therefore, the Scene Flow dataset \cite{Mayer2016A} is used to investigate different aspects of our method in Sec. \ref{sec:exp1}. The KITTI dataset includes two subsets, i.e., KITTI 2012 and KITTI 2015. The KITTI 2012 dataset consists of 194 training image pairs and 195 test image pairs, while the KITTI 2015 dataset consists of 200 training image pairs and 200 test image pairs. These images were recorded in real scenes under different weather conditions. Our method is further compared to the state-of-the-art methods on the KITTI dataset (Sec. \ref{sec:exp2}), with the best results being achieved.

Our method was implemented in CAFFE \cite{jia2014caffe}. All models were optimized using the Adam method  \cite{Kingma2014Adam} with $\beta_1 = 0.9$, $\beta_2$ = 0.999, and a batch size of 2. ``Multi-step" learning rate was used for the training. Specifically, for training on the Scene Flow dataset, the learning rate was initially set to $10^{-4}$ and then reduced by a half at the 20k-th, 35k-th and 50k-th iterations, the training was stopped at the 65k-th iteration. This training procedure was repeated for an additional round to further optimize the model. For fine-tuning on the KITTI dataset, the learning rate was set to $2 \times 10^{-5}$ for the first 20k iterations and then reduced to $10^{-5}$ for the subsequent 120k iterations.

Data augmentation was also conducted for training, including spatial and chromatic transformations. The spatial transformations include rotation, translation, cropping and scaling, while the chromatic transformations includes color, contrast and brightness transformations. This data augmentation can help to learn a robust model against illumination changes and noise.

\subsection{Ablation Experiments}
\label{sec:exp1}

In this section, we present several ablation experiments on the Scene Flow dataset to justify our design choices. For evaluation, we use the end-point-error (EPE), which is calculated as the average euclidean distance between estimated and groundtruth disparity. We also use the percentage of disparities with their EPE larger than $t$ pixels ($> t$ px).

\subsubsection{Multi-scale Skip Connection\label{sec:Multi-scale-Skip-Connection}}

In Sec. \ref{sec:app}, multi-scale skip connection is used to introduce features from different levels to improve disparity estimation and refinement performance. To demonstrate its effectiveness, the multi-scale skip connection scheme of our network  was replaced by a single-scale skip connection scheme, the comparative results are shown in Table \ref{table:ms-ss}. It can be observed that the multi-scale skip connection scheme outperforms its single-scale counterpart, with the EPE being reduced from 2.55 to 2.50. That is because, the output of the first convolution layer contains high frequency information, it produces high reconstruction error for both regions along object boundaries and regions with large color changes. Note that, regions on an object surface far from boundaries usually have a very accurate initial disparity estimation (i.e., the true reconstruction error is small), although large color changes occur due to texture variation. Therefore, the reconstruction errors given by the first convolution layer for these regions are inaccurate. In this case, multi-scale skip connection is able to improve the reliability of resulted reconstruction errors.
Besides, introducing high-level features is also useful for feature constancy calculation, as higher-level features leverage more context information with a wide field of view.

\subsubsection{Feature Constancy for Disparity Refinement}

\begin{table*}[!tb]
\caption{Comparative results on the Scene Flow dataset for networks with different settings on the disparity refinement sub-network. \textcolor{black}{DES-net and DRS-net represent the initial disparity estimation sub-network and the disparity refinement sub-network, respectively.}}
\label{table:bayes}
\begin{center}
\newsavebox{\tablebox}
\begin{lrbox}{\tablebox}
\begin{tabular}{l|c|c|c|c|c|c}
\hline
Model & $>$ 1px & $>$ 3px & $>$5px & EPE & Params. & Time (ms)\\
\hline\hline
DES-net (without disparity refinement)&16.78 & 6.49       &4.31 & 2.81 & \textbf{42.76M} & \textbf{61}\\
\hline
\hline
DES-net + DRS-net without \textcolor{black}{feature correlation $fc$ and $re$} & 15.65     &6.12    &4.11  & 2.72 & 43.30M & 100\\
DES-net + DRS-net without \textcolor{black}{reconstruction error $re$}      & 15.54     &6.10    &4.10  & 2.70 & 43.34M& 111\\
DES-net + DRS-net without \textcolor{black}{feature correlation $fc$}                      & 11.13     &5.32    & 3.79 & 2.56 & 43.31M& 103\\
DES-net + DRS-net without \textcolor{black}{initial disparity $disp_i$ }                                   & 11.64     &5.37    & 3.81 & 2.61 & 43.34M& 114\\
DES-net + DRS-net (iResNet)                                                                                & 10.24     &4.93    & 3.54 & 2.50 & 43.34M& 114\\
\hline
\hline
Refinement $\times$ 2 (iResNet-i2) &9.42    & 4.64  & 3.37 & 2.46 & 43.34M& 131\\
Refinement $\times$ 3 (iResNet-i3) & \textbf{9.28}    & \textbf{4.57}  & \textbf{3.32}  & \textbf{2.45} & 43.34M & 148\\
\hline
\end{tabular}
\end{lrbox}
\scalebox{0.92}{\usebox{\tablebox}}
\end{center}
\end{table*}

\begin{table}[!]
\caption{Comparative results on the Scene Flow dataset for networks with single-scale and multi-scale skip connection.}
\label{table:ms-ss}
\small
\begin{center}
\begin{tabular}{l|c|c|c|c}
\hline
Model & $>$ 1px & $>$ 3px & $>$5px & EPE \\
\hline
Single-scale & 10.90     &5.23    & 3.74   &2.55 \\
Multi-scale   & \textbf{10.24}     &\textbf{4.93}    & \textbf{3.54}   &\textbf{2.50}  \\
\hline
\end{tabular}
\end{center}
\end{table}

\begin{table}[!]
\caption{EPE results on Scene Flow dataset achieved by the proposed iResNet method and the CRL method.}
\label{table:crl}
\small
\centering%
\begin{tabular}{c|c|c|c}
\hline
Method  & EPE & Params.  & Run time(ms) \\
\hline
CRL \cite{Pang2017Cascade}   & 1.60 & 78.77M &  162 \\
iResNet  & \textbf{1.40} & \textbf{43.11M} &  \textbf{90}  \\
\hline
\end{tabular}
\medskip
\end{table}

To seamlessly integrate the \textcolor{black}{initial disparity estimation sub-network (DES-net) and the disparity refinement sub-network (DRS-net)} into a whole network, the feature constancy calculation and its subsequent sub-network play an important role. To demonstrate its effectiveness, we first removed all feature constancy used in our network (as shown in Table \ref{table:iresnet}) and then retrained the model.
The results are shown in Table \ref{table:bayes}. It can be observed that if no feature constancy is introduced for disparity refinement, the performance improvement is very small, with EPE being reduced from 2.81 to 2.72. 

Then, we evaluate the importance of the three information in Eq. (\ref{eq:disp-ftr}), i.e., initial disparity $disp_i$, \textcolor{black}{feature correlation $fc$, and reconstruction error $re$ produced} by the initial disparity, as explained in Sec. \ref{sec:app}. The results are shown in Table \ref{table:bayes}. It is observed that, \textcolor{black}{the reconstruction error $re$} plays the major role for the performance improvement. If the reconstruction error is removed, EPE is increased from 2.50 to 2.70. That is because this term provides the some knowledge about the initial disparity. That is, regions with poor initial disparity can be identified and then be contrapuntally refined. Besides, removing initial disparity $disp_i$ or \textcolor{black}{feature correlation $fc$} from the disparity refinement sub-network slightly degrades the overall performance. Their EPE values are increased from 2.50 to 2.56 and 2.61, respectively. If all the three parts are incorporated, \textcolor{black}{the disparity refinement network can achieve the best performance}.

\begin{figure}[!ht]
  \small
%
\begin{center}
\begin{minipage}[!ht]{0.325\linewidth}
  \centering
  \centerline{\includegraphics[width=2.6 cm]{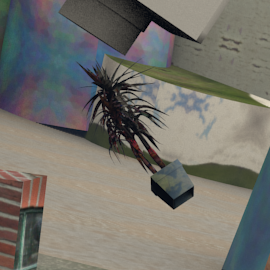}}
\end{minipage}
\begin{minipage}[!ht]{0.325\linewidth}
  \centering
  \centerline{\includegraphics[width=2.6 cm]{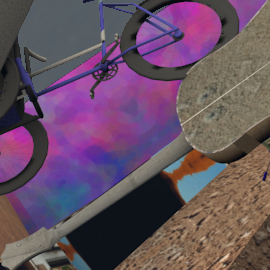}}
\end{minipage}
\begin{minipage}[!ht]{0.325\linewidth}
  \centering
  \centerline{\includegraphics[width=2.6 cm]{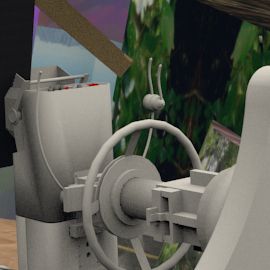}}
\end{minipage}
\\
\begin{minipage}[!ht]{0.325\linewidth}
  \centering
  \centerline{\includegraphics[width=2.6 cm]{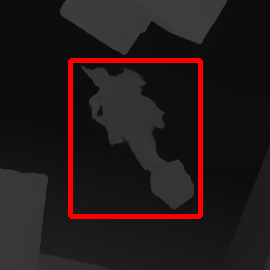}}
\end{minipage}
\begin{minipage}[!ht]{0.325\linewidth}
  \centering
  \centerline{\includegraphics[width=2.6 cm]{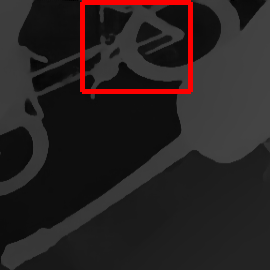}}
\end{minipage}
\begin{minipage}[!ht]{0.325\linewidth}
  \centering
  \centerline{\includegraphics[width=2.6 cm]{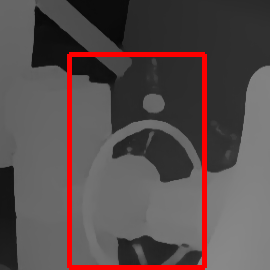}}
\end{minipage}
\\
\begin{minipage}[!ht]{0.325\linewidth}
  \centering
  \centerline{\includegraphics[width=2.6 cm]{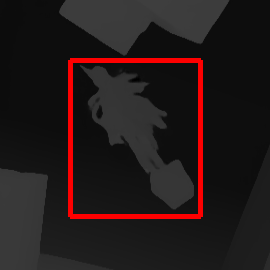}}
\end{minipage}
\begin{minipage}[!ht]{0.325\linewidth}
  \centering
  \centerline{\includegraphics[width=2.6 cm]{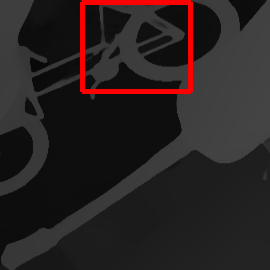}}
\end{minipage}
\begin{minipage}[!ht]{0.325\linewidth}
  \centering
  \centerline{\includegraphics[width=2.6 cm]{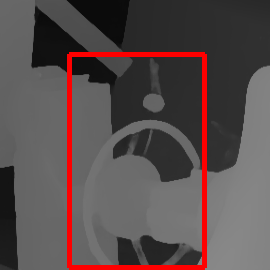}}
\end{minipage}
\\
\begin{minipage}[!ht]{0.325\linewidth}
  \centering
  \centerline{\includegraphics[width=2.6 cm]{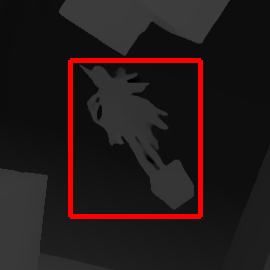}}
\end{minipage}
\begin{minipage}[!ht]{0.325\linewidth}
  \centering
  \centerline{\includegraphics[width=2.6 cm]{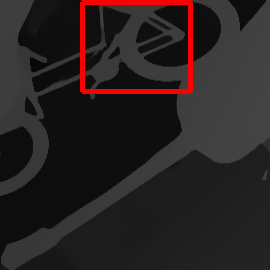}}
\end{minipage}
\begin{minipage}[!ht]{0.325\linewidth}
  \centering
  \centerline{\includegraphics[width=2.6 cm]{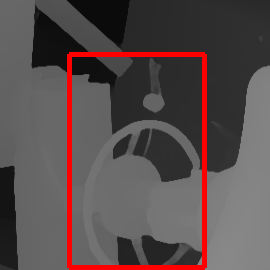}}
\end{minipage}
\\
\begin{minipage}[!ht]{0.325\linewidth}
  \centering
  \centerline{\includegraphics[width=2.6 cm]{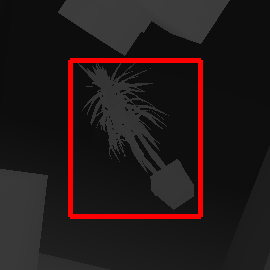}}
\end{minipage}
\begin{minipage}[!ht]{0.325\linewidth}
  \centering
  \centerline{\includegraphics[width=2.6 cm]{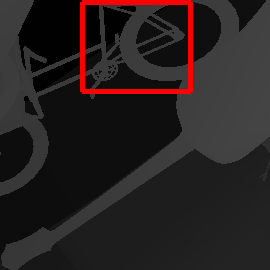}}
\end{minipage}
\begin{minipage}[!ht]{0.325\linewidth}
  \centering
  \centerline{\includegraphics[width=2.6 cm]{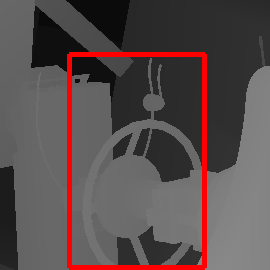}}
\end{minipage}
\\
\caption{Disparity refinement results on the Scene Flow testing set under different iterations. The first \textcolor{black}{row} represents the input images, the second \textcolor{black}{row} shows the initial disparity without any refinement, the third and fourth \textcolor{black}{rows} show the refined disparity after 1 and 2 iterations, respectively. The last \textcolor{black}{row} gives the groundtruth disparity.}
\label{fig:sceneflow}
\end{center}
\end{figure}

\begin{figure}[!htb]
\begin{minipage}[!htb]{0.02\linewidth}
\scriptsize
  \centering
  {\rotatebox{90}{Inputs}}
\end{minipage}
\begin{minipage}[!htb]{0.48\linewidth}
  \centering
  \centerline{\includegraphics[width=4.0 cm]{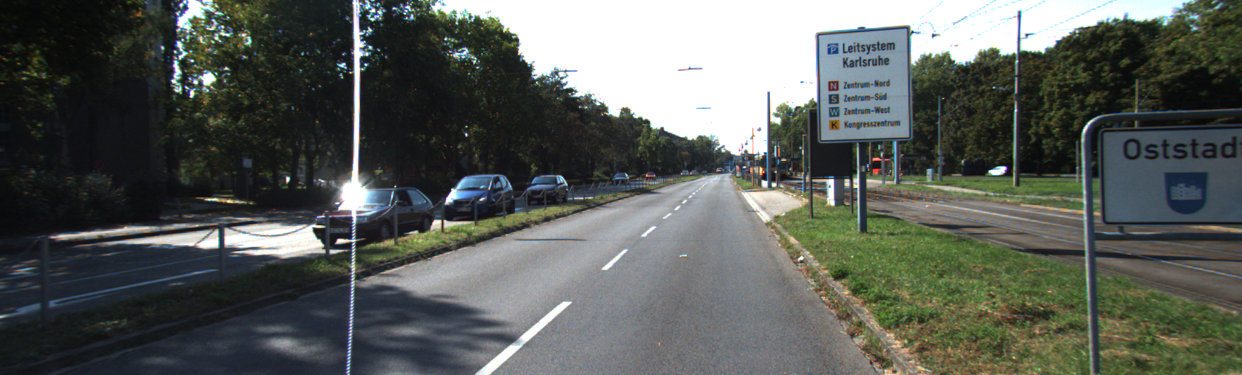}}
\end{minipage}
%
\begin{minipage}[!htb]{0.48\linewidth}
  \centering
  \centerline{\includegraphics[width=4.0 cm]{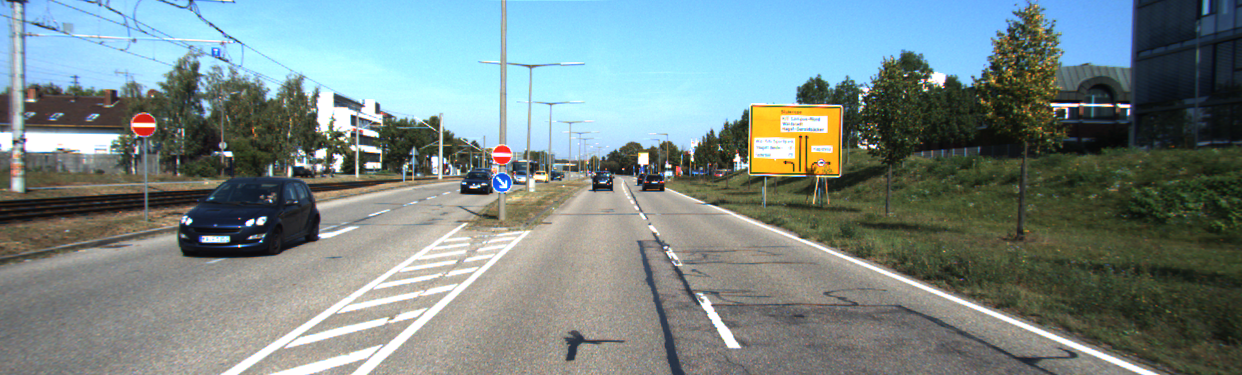}}
\end{minipage}
\\
\begin{minipage}[!htb]{0.02\linewidth}
\scriptsize
  \centering
  {\rotatebox{90}{DES-net}}
\end{minipage}
\begin{minipage}[!htb]{0.48\linewidth}
  \centering
  \centerline{\includegraphics[width=4.0 cm]{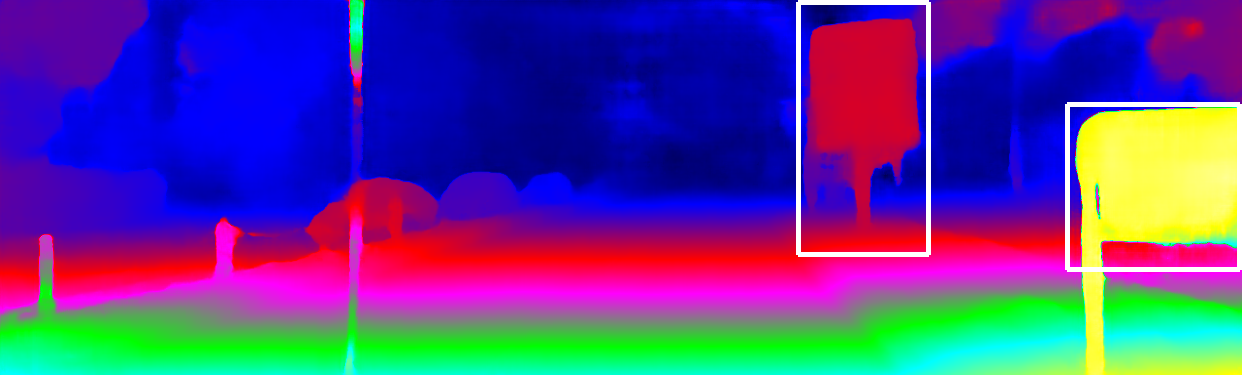}}
\end{minipage}
\begin{minipage}[!htb]{0.48\linewidth}
  \centering
  \centerline{\includegraphics[width=4.0 cm]{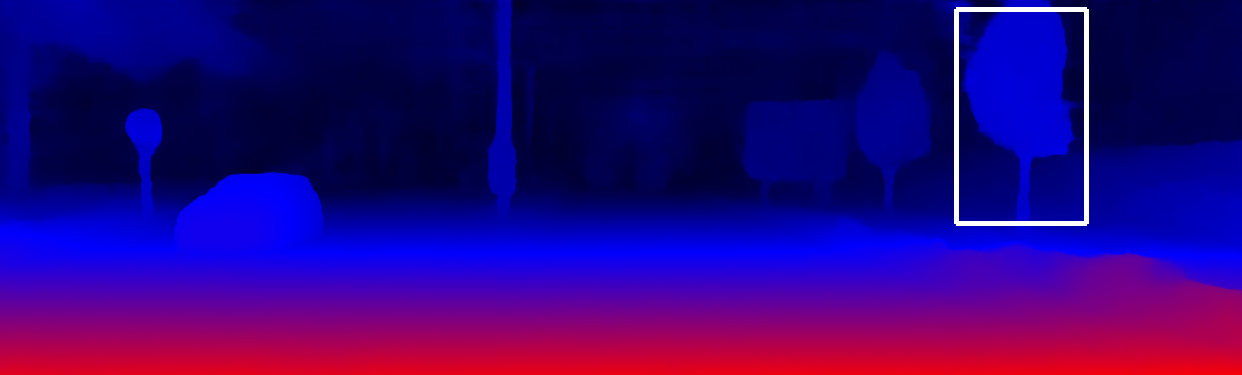}}
\end{minipage}
\\
\begin{minipage}[!htb]{0.02\linewidth}
\scriptsize
  \centering
  {\rotatebox{90}{iResNet-i2}}
\end{minipage}
\begin{minipage}[!htb]{0.48\linewidth}
  \centering
  \centerline{\includegraphics[width=4.0 cm]{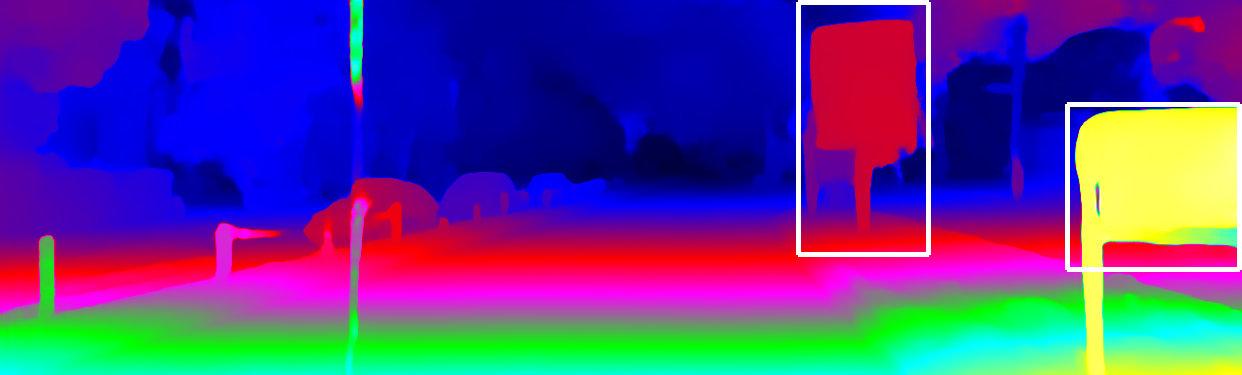}}
\end{minipage}
%
\begin{minipage}[!htb]{0.48\linewidth}
  \centering
  \centerline{\includegraphics[width=4.0 cm]{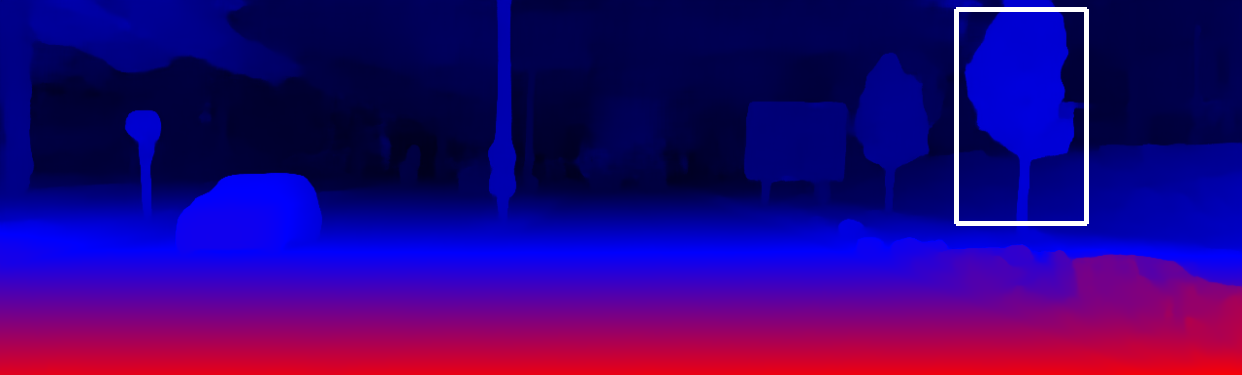}}
\end{minipage}
\\
\begin{minipage}[!htb]{0.02\linewidth}
\scriptsize
  \centering
  {\rotatebox{90}{CRL}}
\end{minipage}
\begin{minipage}[!htb]{0.48\linewidth}
  \centering
  \centerline{\includegraphics[width=4.0 cm]{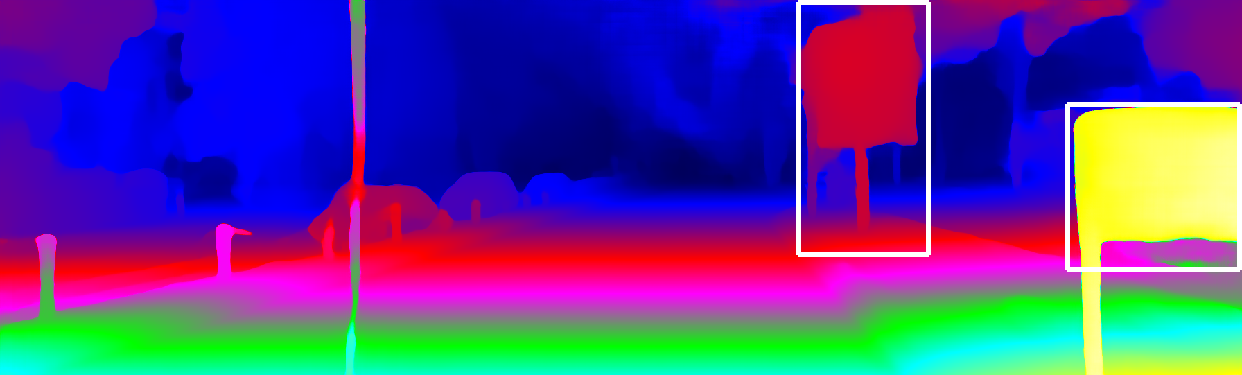}}
\end{minipage}
%
\begin{minipage}[!htb]{0.48\linewidth}
  \centering
  \centerline{\includegraphics[width=4.0 cm]{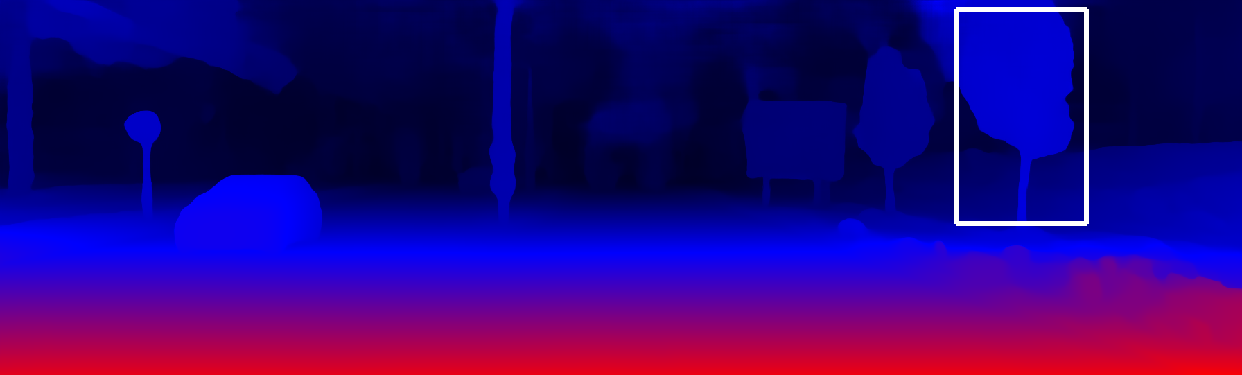}}
\end{minipage}
\\
\begin{minipage}[!htb]{0.02\linewidth}
\scriptsize
  \centering
  {\rotatebox{90}{GC-NET}}
\end{minipage}
\begin{minipage}[!htb]{0.48\linewidth}
  \centering
  \centerline{\includegraphics[width=4.0 cm]{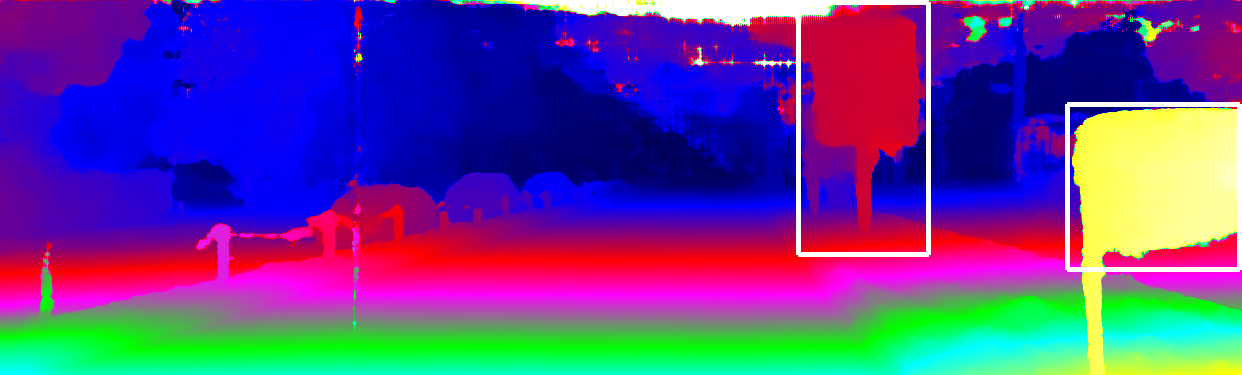}}
\end{minipage}
%
\begin{minipage}[!htb]{0.48\linewidth}
  \centering
  \centerline{\includegraphics[width=4.0 cm]{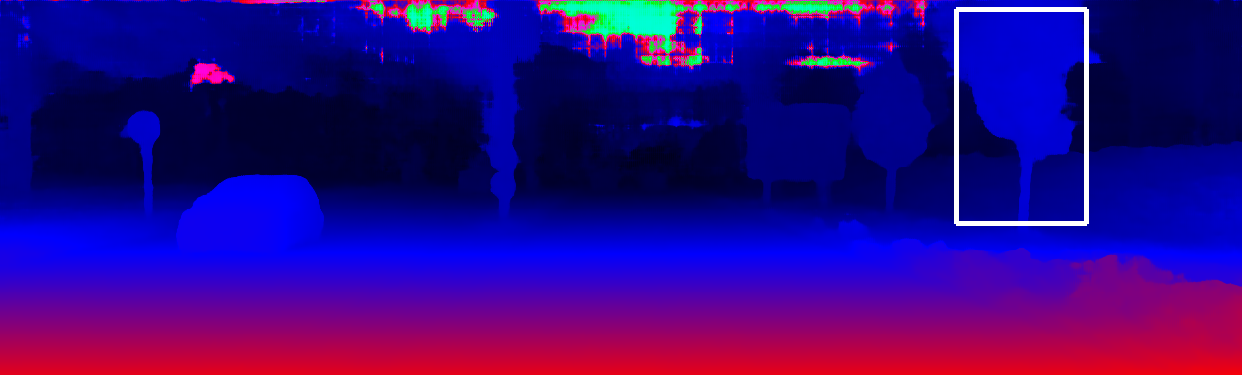}}
\end{minipage}
\\
\begin{minipage}[!htb]{0.02\linewidth}
\scriptsize
  \centering
  {\rotatebox{90}{LResMatch}}
\end{minipage}
\begin{minipage}[!htb]{0.48\linewidth}
  \centering
  \centerline{\includegraphics[width=4.0 cm]{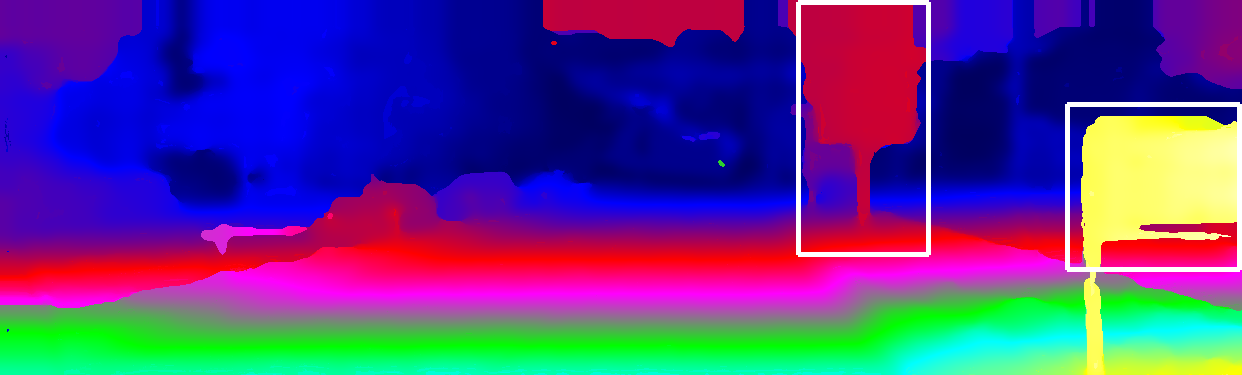}}
\end{minipage}
%
\begin{minipage}[!htb]{0.48\linewidth}
  \centering
  \centerline{\includegraphics[width=4.0 cm]{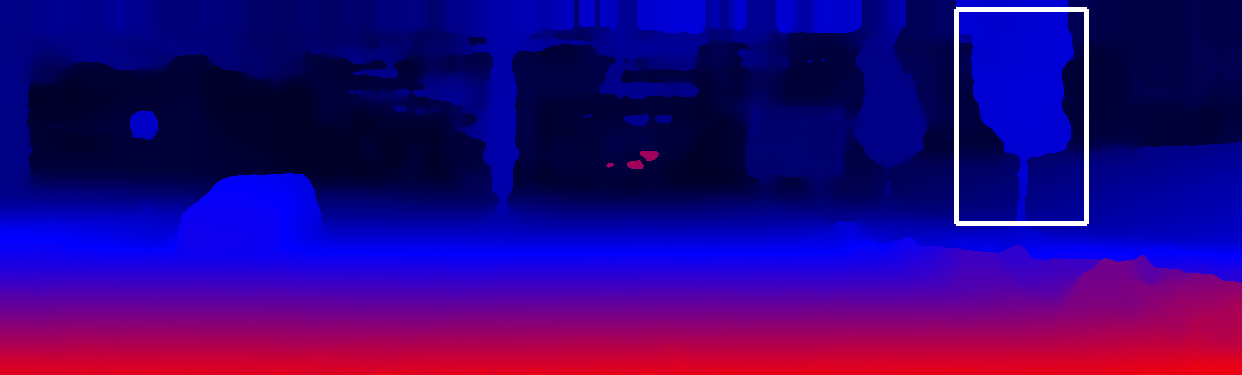}}
\end{minipage}
\\
\begin{minipage}[!htb]{0.02\linewidth}
\scriptsize
  \centering
  {\rotatebox{90}{MC-CNN-acrt}}
\end{minipage}
\begin{minipage}[!htb]{0.48\linewidth}
  \centering
  \centerline{\includegraphics[width=4.0 cm]{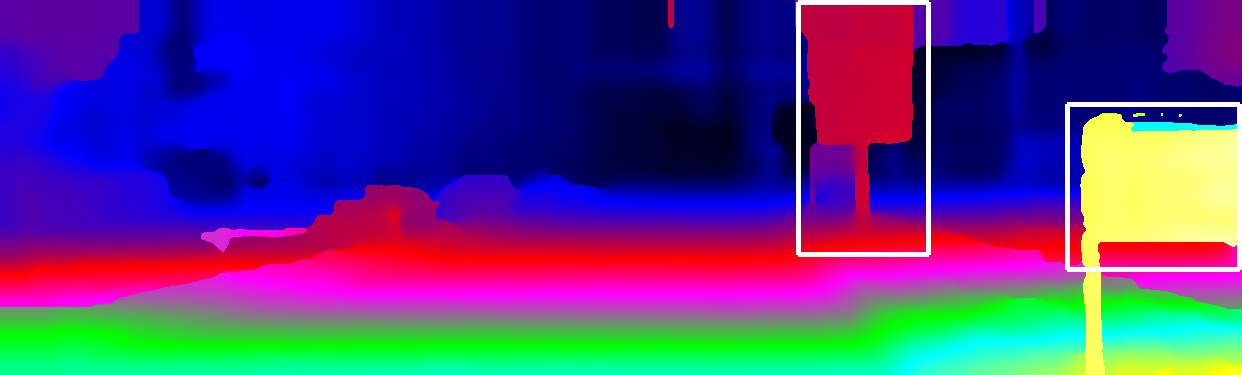}}
\end{minipage}
%
\begin{minipage}[!htb]{0.48\linewidth}
  \centering
  \centerline{\includegraphics[width=4.0 cm]{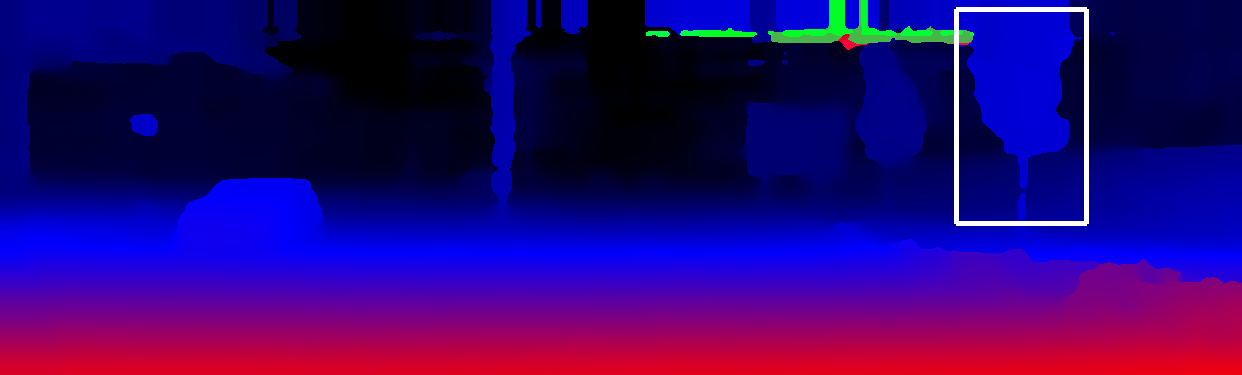}}
\end{minipage}
\\
\begin{minipage}[!htb]{0.02\linewidth}
\scriptsize
  \centering
  {\rotatebox{90}{DispNetC}}
\end{minipage}
\begin{minipage}[!htb]{0.48\linewidth}
  \centering
  \centerline{\includegraphics[width=4.0 cm]{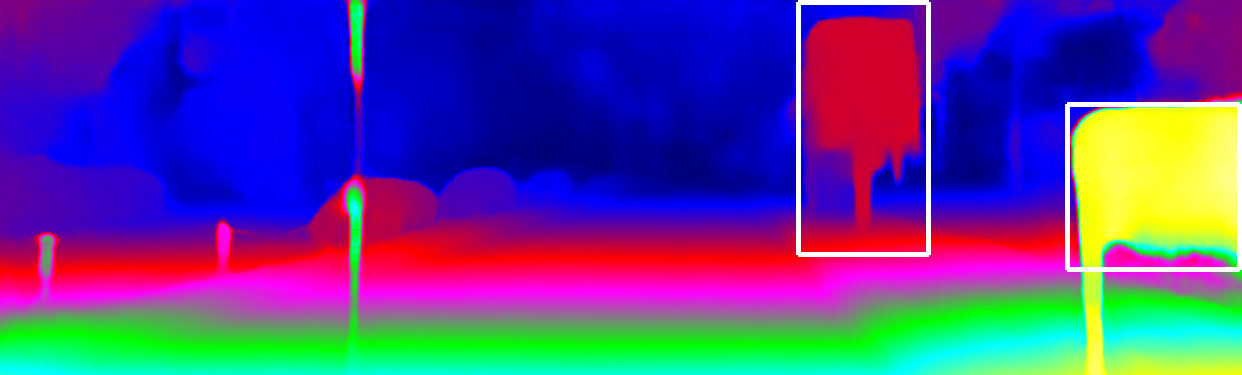}}
\end{minipage}
%
\begin{minipage}[!htb]{0.48\linewidth}
  \centering
  \centerline{\includegraphics[width=4.0 cm]{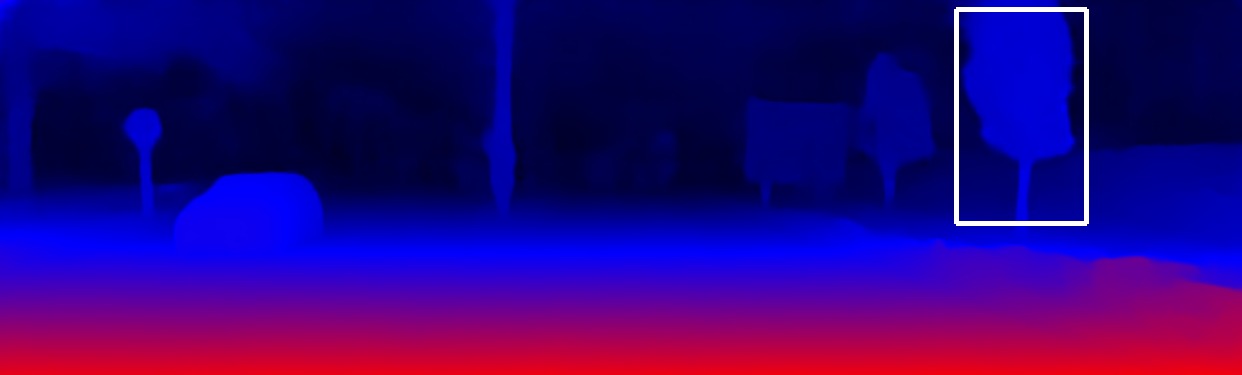}}
\end{minipage}
\\
\caption{Comparison with other state-of-the-art methods on the KITTI 2015 dataset. The images in the first row are input images from KITTI 2015. Our iResNet-i2 refinement results (the third row) can greatly improve the initial disparity (the second row), and give better visualization effect than other methods, especially in the upper part of images, where there is no groundtruth in these region.}
\label{fig:kitti2015}
\end{figure}

\subsubsection{Iterative Refinement}






Iterative feature constancy calculation can further improve the overall performance. In practice, we do not train multiple DRS-nets. Instead, we directly stack another DRS-net, whose weights are exactly the same as the original DRS-net, at the top of the whole network.
As the number of iterations is increased, the performance improvement is decreased rapidly. Specifically, the first iteration can reduce EPE from 2.50 to 2.46, while the second iteration can only reduce EPE from 2.46 to 2.45. Moreover, there is no obvious performance improvement after the third iteration of refinement. It can be concluded that: 1) Our framework can efficiently extract useful information for disparity estimation using feature constancy information; 2) The information contained in feature space is still limited. Therefore, it is possible to improve the disparity refinement performance by introducing more powerful features.

To further demonstrate the effectiveness of iterative refinement, the disparity estimation results for 2 iterations are shown in Fig. \ref{fig:sceneflow}. It can be observed that, lots of details cannot be accurately estimated in the initial disparity, e.g., the areas shown in red rectangles. However, after two iterations of refinement, most inaccurate estimations can be corrected, and the refined disparity map looks more smooth.

\begin{table*}[]
\caption{Results on the KITTI 2012 dataset.}\label{table:kitti2012}
\begin{center}
\begin{lrbox}{\tablebox}
\begin{tabular}{l|cc|cc|cc|cc|c}
\hline
 &        \multicolumn{2}{c|}{$>$ 2px}         & \multicolumn{2}{c|}{$>$ 3px} & \multicolumn{2}{c|}{$>$ 4px} & \multicolumn{2}{c|}{$>$ 5px}  & Runtime\\
 &Out-Noc&Out-All&Out-Noc&Out-All&Out-Noc&Out-All&Out-Noc&Out-All&(s)\\
\hline
GC-NET \cite{Kendall2017End}       &2.71  &3.46         &1.77  &2.30      &1.36 & 1.77   &1.12 &1.46 &0.9\\
L-ResMatch \cite{Shaked2016Improved} &3.64  &5.06         &2.27 &3.40      &1.76 &2.67     &1.50 &2.26 &48\\
SGM-Net \cite{SekiSGMNetsSM}     &3.60  &5.15         &2.29  &3.50     &1.83 &2.80    &1.60 & 2.36 & 67 \\
SsSMnet \cite{Zhong2017SelfSupervisedLF}    &3.34  &4.24         & 2.30 &3.00      &1.82 & 2.39  &1.53 & 2.01& 0.8 \\
PBCP \cite{Seki2016PatchBC}         &3.62  &5.01        &2.36 & 3.45     &1.88 &2.28    &1.62 & 2.32& 68\\
Displets v2 \cite{Gney2015DispletsRS}  &3.43 &4.46         &2.37&3.09       &1.97&2.52    &1.72&2.17& 265\\
MC-CNN-acrt \cite{zbontar2016stereo}     &3.90 &5.45         &2.43&3.63     &1.90&2.85 &1.64&2.39& 67\\
\hline
DES-net (ours) &4.88  	&5.54  &2.66  	&3.12  	&1.78  	&2.11  	&1.33  	&1.59 & \textbf{0.05} \\
iResNet-i2 (ours)&\textbf{2.69}&\textbf{3.34}&\textbf{1.71}&\textbf{2.16}&\textbf{1.30}&\textbf{1.63}&\textbf{1.06}&\textbf{1.32}&0.12 \\
\hline
\end{tabular}
\end{lrbox}
\scalebox{0.81}{\usebox{\tablebox}}
\end{center}
\end{table*}

\subsubsection{\textcolor{black}{Feature Constancy \textit{vs} Color Constancy}}
\textcolor{black}{
In this experiment, the superiority of feature constancy for disparity refinement over color constancy is demonstrated. Our method was compared to the cascade residual learning (CRL) method \cite{Pang2017Cascade}, which calculated the reconstruction error in the color space. Intuitively, calculating the reconstruction error in the feature space could obtain more robust results, since the learned features are less sensetive to noise and luminance changes. Besides, by sharing the features with the first network, the second network can be designed shadower, which would improve the feature effections, and reduce running time.}
CRL used one network (i.e., DispNetC) for disparity calculation and an additional network (i.e., DispResNet) for disparity refinement. For fair comparison, we also use DispNetC \cite{Mayer2016A} without any fine-tuning as our disparity estimation sub-network.
Note that, in their experiments on the Scene Flow dataset, disparity images (and their corresponding stereo pairs) with more than 25\% of their disparity
values greater than 300 were removed. We followed the same protocol in this comparison.
Comparative results are shown in Table \ref{table:crl}. The EPE result of CRL is taken from the paper \cite{Pang2017Cascade}, and its run time was tested on Nvidia Titan X (Pascal) using our implementation.
It can be seen that our method (\textcolor{black}{using feature constancy}) significantly outperforms CRL (\textcolor{black}{using color constancy}). The EPE values achieved by our iResNet method and the CRL method are 1.40 and 1.60, respectively. Moreover, our method also requires fewer parameters and costs less computational time.

\subsection{Benchmark Results}
\label{sec:exp2}

We further compared our method to several existing methods on the  KITTI 2012 and KITTI 2015 benchmarks, including GC-NET \cite{Kendall2017End}, L-ResMatch \cite{Shaked2016Improved}, SGM-Net \cite{SekiSGMNetsSM}, SsSMnet \cite{Zhong2017SelfSupervisedLF}, PBCP \cite{Seki2016PatchBC}, Displets \cite{Gney2015DispletsRS}, and MC-CNN \cite{zbontar2016stereo}.

For the evaluation on KITTI 2012, we used the percentage of erroneous pixels in non-occluded (Out-Noc) and all (Out-All) areas. Here a pixel is considered to be erroneous if its disparity EPE is larger than $t$ pixels ( $> t$ px).
For the evaluation on KITTI 2015, we used the percentage of erroneous pixels in background ($D1$-$bg$), foreground ($D1$-$fg$) or all pixels ($D1$-$all$) in the non-occluded or all regions. Here, a pixel is considered to be \textcolor{black}{correct if its disparity EPE is less than 3 or 5$\%$ pixels}.

The results are shown in Tables \ref{table:kitti2012} and \ref{table:kitti2015}. For our method, the results for both of the basic model (without disparity refinement) and the final iResNet model (with disparity refinement of 2 iterations) are presented. It is clear that the disparity refinement sub-network can consistently improve the performance by a notable margin.
 Moreover, our iResNet model achieves the best disparity estimation performance on both the KITTI 2012 and KITTI 2015 datasets in different scenarios. Note that, our method is also highly efficient, it achieves the shortest run time on the KITTI 2012 dataset. The overall run time tested on a single Nvidia Titan X (Pascal) GPU is only 0.12s.

Figure \ref{fig:kitti2015} illustrates few qualitative results achieved by our method and several state-of-the-art methods on the KITTI 2015 dataset. It can be observed that our method produces more smooth disparity estimation results, \textcolor{black}{as shown in the rectangle marked in Figure \ref{fig:kitti2015}. On one hand, our disparity refinement sub-network can effectively improve the quality of the initial disparity estimated by DES-net, with many details being recovered. On the other hand, }our method gives much better results than other methods in the upper parts of these images \textcolor{black}{as denoted in red rectangles in Fig. \ref{fig:kitti2015}}. In fact, the upper parts of these images correspond to sky or regions beyond the working distance of a lidar. In that case, groundtruth disparity cannot be provided for these regions for training, making the learning based methods to deteriorate. From Fig. \ref{fig:kitti2015}, we can see that \textcolor{black}{the performance of other methods in these regions is relatively poor. However, our method can still provide acceptable results, with more accurate disparity estimation along boundaries. This also indicates that our method generalizes well on unseen data. 
To further demonstrate the generalization capability of our method, the iResNet-i2, CRL \cite{Pang2017Cascade} and DispNetC \cite{Mayer2016A} models trained on the KITTI 2015 training set are tested on the KITTI 2015 and 2012 test sets without fine-tuning. The achieved \textit{D}1-\textit{all} errors are shown in Table \ref{table:gen_to_2012} in this page. It is clear that our method achieves the smallest performance drop.}


\begin{table}[]
\caption{Results on the KITTI 2015 dataset.}
\label{table:kitti2015}
\begin{center}
\begin{lrbox}{\tablebox}
\begin{tabular}{p{2.5cm}|p{0.9cm}p{0.9cm}p{0.9cm}|p{0.9cm}p{0.9cm}p{0.9cm}}
\hline
 &        \multicolumn{3}{c|}{All Pixels}         & \multicolumn{3}{c}{Non-Occluded Pixels}\\
 & D1-bg & D1-fg & D1-all & D1-bg & D1-fg & D1-all \\
\hline
CRL \cite{Pang2017Cascade}&2.48&3.59&2.67&2.32&3.12&2.45\\
GC-NET \cite{Kendall2017End}&2.21 &6.16 &2.87&2.02&5.58&2.61\\
DRR \cite{Gidaris2017Detect} & 2.58&6.04&3.16&2.34&4.87&2.76\\
SsSMnet \cite{Zhong2017SelfSupervisedLF} &2.70&6.92&3.40&2.46&6.13&3.06\\
L-ResMatch \cite{Shaked2016Improved}&2.72&6.95&3.42&2.35&5.74&2.91\\
Displets v2 \cite{Gney2015DispletsRS}&3.00&5.56&3.43&2.73&4.95&3.09\\
SGM-Net \cite{SekiSGMNetsSM} & 2.66&8.64&3.66&2.23&7.44&3.09\\
MC-CNN \cite{zbontar2016stereo} & 2.89&8.88&3.88&2.48&7.64&3.33\\
DispNetC \cite{Mayer2016A}& 4.32 &4.41 &4.34&4.11&3.72&4.05\\
\hline
DES-net (ours)&3.13&3.87&3.25&2.94&3.21&2.98\\
iResNet-i2 (ours)&\textbf{2.25}&\textbf{3.40}&\textbf{2.44}&\textbf{2.07}&\textbf{2.76}&\textbf{2.19}\\
\hline
\end{tabular}
\end{lrbox}
\scalebox{0.77}{\usebox{\tablebox}}
\end{center}
\end{table}

\begin{table}[]
\caption{The generalization performance from KITTI 2015 to KITTI 2012 achieved by three methods.}
\label{table:gen_to_2012}
\centering
\begin{tabular}{c|c|c|c}
\hline
\footnotesize Methods &       \footnotesize KITTI 2015      & \footnotesize KITTI 2012 & \footnotesize Performance Drop \\
\hline
\footnotesize iResNet-i2 (ours)&\footnotesize \textbf{2.44}& \footnotesize \textbf{3.62}& \footnotesize \textbf{1.18}\\
\footnotesize CRL \cite{Pang2017Cascade}& \footnotesize 2.67& \footnotesize 4.82& \footnotesize 2.15\\
\footnotesize DispNetC \cite{Mayer2016A}& \footnotesize 4.34& \footnotesize 9.64& \footnotesize 5.3\\
\hline
\end{tabular}
\vspace{-8pt}
\end{table}

\section{Conclusion}
In this work, we propose a network architecture to integrate the four steps of stereo matching for joint training. Our network first estimates an initial disparity, and then \textcolor{black}{uses two feature constancy terms to refine the disparity. The refinement is performed using both feature correlation and reconstruction error}, which makes the network easy for optimization. Experimental results show that the proposed method achieves the state-of-the-art disparity estimation performance on the KITTI 2012 and KITTI 2015 datasets. Moreover, our method is also highly efficient for calculation.

\end{document}